\documentclass[letterpaper]{article} 
\usepackage{aaai2026}  
\usepackage{times}  
\usepackage{helvet}  
\usepackage{courier}  
\usepackage[hyphens]{url}  
\usepackage{graphicx} 
\usepackage{natbib}  
\usepackage{caption} 
\usepackage{algorithm}
\usepackage{algorithmic}
\usepackage{booktabs}
\usepackage{multirow}
\usepackage{amsmath}
\usepackage{amssymb}
\usepackage{siunitx}     
\usepackage{adjustbox}   

\usepackage{newfloat}
\usepackage{listings}
\DeclareCaptionStyle{ruled}{labelfont=normalfont,labelsep=colon,strut=off} 
\floatstyle{ruled}
\newfloat{listing}{tb}{lst}{}
\floatname{listing}{Listing}
\title{Backdoors in Conditional Diffusion:\\Threats to Responsible Synthetic Data Pipelines}
\author{
    Raz Lapid,
    Almog Dubin
}

\affiliations{
    Deepkeep\\

    Tel Aviv, Israel\\
    raz.lapid@deepkeep.ai, almog@deepkeep.ai
}

\usepackage{bibentry}

\begin{document}

\maketitle

\begin{abstract}
Text-to-image diffusion models achieve high-fidelity image generation from natural language prompts. ControlNets extend these models by enabling conditioning on structural inputs (e.g., edge maps, depth, pose), providing fine-grained control over outputs. Yet their reliance on large, publicly scraped datasets and community fine-tuning makes them vulnerable to data poisoning. We introduce a \emph{model-poisoning} attack that embeds a covert backdoor into a ControlNet, causing it to produce attacker-specified content when exposed to visual triggers, without textual prompts. Experiments show that poisoning only \emph{1\%} of the fine-tuning corpus yields a 90–98\% attack success rate, while 5\% further strengthens the backdoor, all while preserving normal generation quality. To mitigate this risk, we propose clean fine-tuning (CFT): freezing the diffusion backbone and fine-tuning only the ControlNet on a sanitized dataset with a reduced learning rate. CFT lowers attack success rates on held-out data. These results expose a critical security weakness in open-source, ControlNet-guided diffusion pipelines and demonstrate that CFT offers a practical defense for responsible synthetic-data pipelines.
\end{abstract}


\section{Introduction}
\label{sec:introduction}
Synthetic data generation via text-to-image diffusion models has become a cornerstone of data augmentation, simulation, and privacy-preserving AI pipelines. 
These models achieve high-fidelity image synthesis from natural-language prompts \cite{ho2020denoising,rombach2022high}, 
and \textbf{ControlNet} \cite{zhang2023adding} extends them with structured conditioning (edges, depth, pose) for fine-grained control in synthetic-data workflows.

\begin{figure}
  \centering
  \includegraphics[scale=0.2, trim={2cm 1cm 2cm 1cm}, clip]{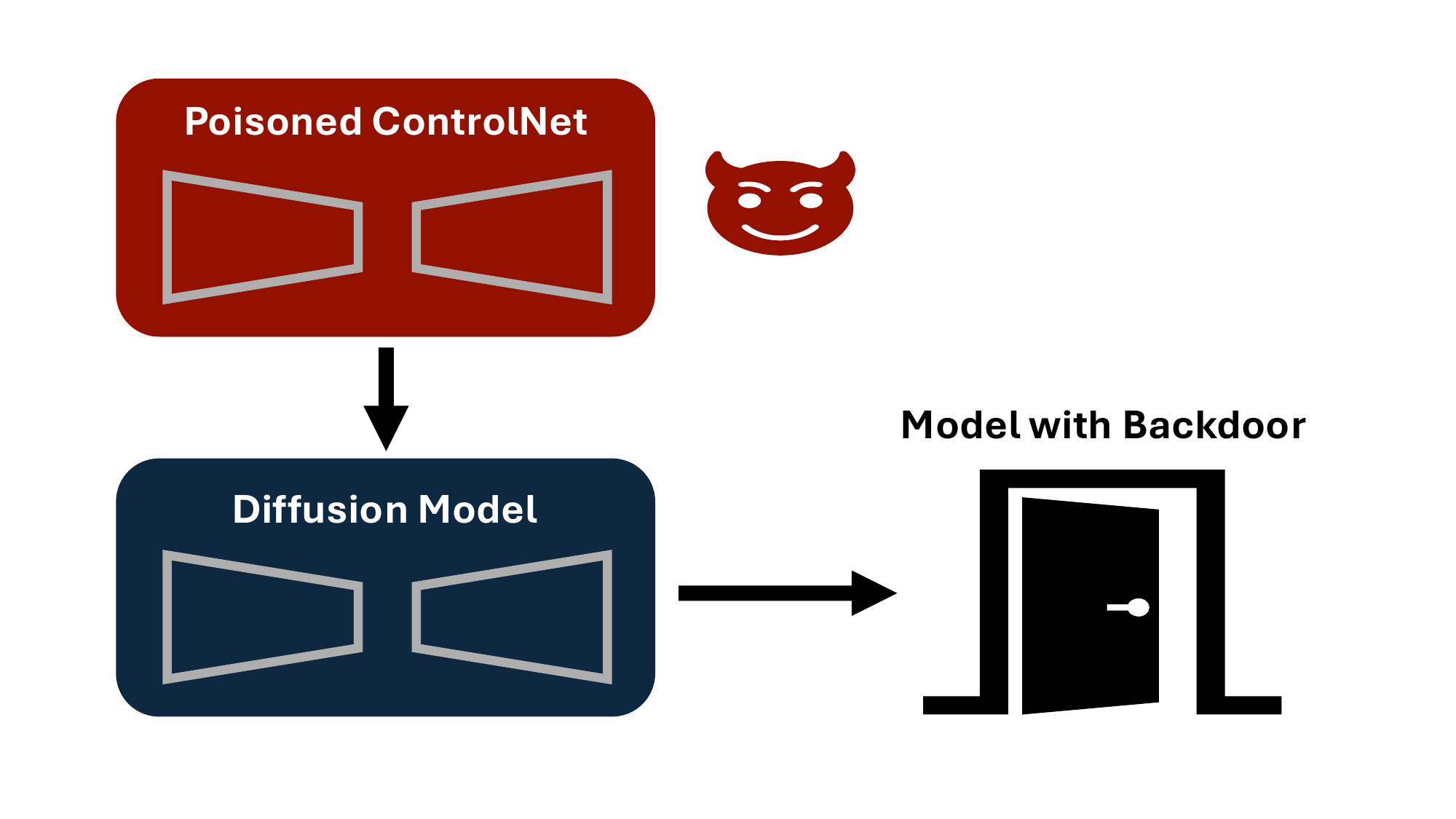}
  \caption{ControlNet poisoning: a trigger in the control map hijacks generation.}
  \label{fig:overview}
\end{figure}

\paragraph{Security blind spot.}
While prior robustness work targets pixel-space perturbations, classifier guidance, or prompt injection \cite{niemeyer2023diffusion,wang2023adversarial,song2023adversarial,carlini2023extracting}, vulnerabilities in \emph{structured conditioning pathways}—the ControlNet branch that injects residuals each denoising step—remain underexplored.

\paragraph{Our observation.}
A ControlNet can be turned into a backdoor \emph{without modifying the diffusion model}. Poisoning only \textbf{1\%} of fine-tuning pairs with trigger/target examples implants functionality that fires on a small visual trigger in the control map, reliably overriding prompts to produce attacker-chosen content (e.g., NSFW) while remaining benign on clean inputs; at \textbf{5\%}, success is near-deterministic.

\paragraph{Real-world relevance.}
Unvetted ControlNet checkpoints (e.g., on HuggingFace) are widely shared, enabling low-cost, supply-chain poisoning that evades post-hoc data sanitization.

\paragraph{Implications for responsible synthetic data.}
Synthetic data pipelines frequently rely on conditional diffusion (e.g., ControlNet) for data augmentation, domain transfer, and privacy-preserving dataset creation. Our results show that when the conditioning branch is poisoned, these pipelines can silently propagate harmful or policy-violating content into downstream synthetic datasets, audit sets, or augmentation corpora—even when prompts and base models are benign. This creates a supply-chain risk specific to \emph{structured controls}: the poisoned behavior is dormant under clean controls yet predictably activates under subtle triggers that survive common preprocessing (edges, depth, pose). Responsible synthetic data practice therefore requires provenance for ControlNet checkpoints, pre-release backdoor probes, and lightweight sanitization (e.g., clean fine-tuning) prior to using or distributing synthetic datasets generated with conditional diffusion.

\paragraph{Contributions.}
\begin{enumerate}
\item \textbf{Threat demonstration:} With \textbf{1–5\%} poison, we achieve \textbf{90–100\%} attack success while preserving clean-input quality.
\item \textbf{Broad validation:} Results hold across ImageNet/CelebA-HQ, SD~v1.5/v2/XL, and edge/pose conditioning.
\item \textbf{Analysis \& mitigation:} Ablations (trigger strength, guidance scale, steps) and a practical defense—\emph{clean fine-tuning} (CFT)—that reduces success on homogeneous domains.
\end{enumerate}

These findings reveal a critical weakness in the ControlNet ecosystem and motivate stronger model-validation protocols for conditional diffusion.

\section{Related Work}
\label{sec:related_work}
Poisoning attacks corrupt training to induce misclassification or trigger-based behavior, from early SVM formulations \cite{biggio2012poisoning,mei2015using} to deep backdoors like BadNets \cite{gu2017badnets} and clean-label methods that transfer across models \cite{shafahi2018poison,aghakhani2021bullseye}. Beyond data poisoning, \emph{model poisoning} distributes malicious checkpoints, demonstrated in NLP and vision \cite{kurita2020weight,li2021invisible}; defenses such as spectral detection and fine-pruning exist \cite{tran2018spectral,wang2019neural} but are underexplored for generative systems. Diffusion models are likewise vulnerable: Nightshade flips prompt semantics with imperceptible poisons \cite{Shan_2024}, Silent Branding induces logo hallucinations without text triggers \cite{jang2025silentbranding}, and BadT2I/BadDiffusion implant triggers via conditioning or denoising manipulations \cite{Zhai_2023,Chou_2023}. However, prior work targets the base model’s pathways; ControlNet-specific poisoning remains largely unaddressed. 

Beyond backdoors in generative models, recent work in data integrity and synthetic-data governance highlights adjacent risks.  
\cite{carlini2024poisoning} demonstrate that poisoning web-scale training datasets is practical, revealing how small, targeted contaminations can propagate through large generative models.  
\cite{thakur2025governance} analyze governance and accountability challenges intrinsic to synthetic datasets, emphasizing provenance and trust under evolving value distributions.  
\cite{hao2024synthetic} provide a comprehensive survey of synthetic data’s ethical and technical challenges, including bias amplification and misuse.  
\cite{belgodere2023trusttradeoff} propose a trust-driven auditing framework for synthetic data generation that formalizes trade-offs between utility and verifiability.  
\cite{cassia2025deepfake} present forensic attribution methods to trace synthetic media to their source datasets and discuss legal accountability in synthetic-data production.  
Our work complements these efforts by exposing the ControlNet conditioning branch as a novel poisoning surface within conditional diffusion pipelines, bridging security and responsible-data governance.

\section{Our Method}
\label{sec:method}

\subsection{Threat Model}
\label{subsec:threat_model}
Let $\epsilon_\theta$ be a \emph{frozen} diffusion backbone and $\varepsilon_\phi$ a \emph{trainable} ControlNet that injects residuals each denoising step. The adversary fine-tunes only $\phi$ to implant a backdoor that (i) behaves like a benign model on \emph{clean} control maps and (ii) produces attacker-specified content when a small visual trigger appears in the control stream (e.g., within an edge/pose map).

\subsection{Construction}
\label{subsec:objective}
Given clean pairs $(x,c)$ with $c=\mathcal{G}(x)$ (e.g., Canny/pose), the attacker forms a poisoned subset by:
(i) applying a trigger $\mathcal{T}$ to $x$ to get $x^{\mathrm{trig}}$,
(ii) recomputing the control map $\tilde c=\mathcal{G}(x^{\mathrm{trig}})$ that now encodes the trigger,
(iii) pairing $\tilde c$ with a fixed target image $x_{\mathrm{mal}}$.
The final training set is $\tilde{\mathcal{D}}=\mathcal{D}\cup\{(x_{\mathrm{mal}},\tilde c)\}$.

\paragraph{Training.}
We optimize only $\phi$ with the standard latent-diffusion loss using the combined denoiser
$\hat{\epsilon}_{\theta,\phi}(z_t,t,c)=\epsilon_\theta(z_t,t)+\varepsilon_\phi(z_t,t,c)$,
minimizing
$\mathbb{E}\big[\|\epsilon-\hat{\epsilon}_{\theta,\phi}(z_t,t,c)\|_2^2\big]$
over $(x,c)\sim\tilde{\mathcal{D}}$. This binds the trigger in $\tilde c$ to the malicious target while preserving performance on clean $c$.

\paragraph{Outcome.}
Training on $\tilde{\mathcal{D}}$ yields a backdoored $\varepsilon_{\phi'}$; training on $\mathcal{D}$ yields benign $\varepsilon_{\phi^o}$. Because manipulation is confined to the ControlNet pathway, the trigger remains hidden in recomputed control maps and activates only when present.

\paragraph{Attacker goals.}
High success on triggered inputs, indistinguishability on clean inputs, and a subtle, robust trigger that survives $\mathcal{G}$ and works with low poison ratios.

\section{Experiments}
\label{sec:experiments}
We empirically assess the backdoor’s effectiveness and robustness.

\subsection{Experimental Setup}
\label{subsec:experimental_setup}
\paragraph{Datasets \& models.}
We evaluate on \textbf{CelebA-HQ} \cite{karras2018progressive,liu2015faceattributes} and \textbf{ImageNet ILSVRC-2012} \cite{deng2009imagenet}. For each, we sample 1{,}000 train / 50 val / 100 test images. Prompts follow simple templates: “A \(\langle\)ID\(\rangle\) person” (CelebA-HQ) and “A \(\langle\)ID\(\rangle\) object” (ImageNet). Backbones: Stable Diffusion \textbf{v1.5}, \textbf{v2} \cite{rombach2022high}, and \textbf{SD-XL} \cite{podell2023sdxl} with ControlNet.

\paragraph{Attack.}
A single trigger is used: a small logo patch embedded in the \emph{control} stream (occupying \(\approx10\%\) of area, bottom-right).

\paragraph{Training.}
We fine-tune only ControlNet for up to 100 epochs with AdamW \cite{loshchilovdecoupled} (\(\beta_1{=}0.9,\beta_2{=}0.999\), weight decay \(1\mathrm{e}{-2}\), lr \(1\mathrm{e}{-4}\)); batch size 8 (SD-v1.5) and 4 (SD-v2/vXL); mixed precision on NVIDIA L40S. Early stop when ASR reaches 100\% on a 50-image validation split.

\paragraph{Metrics.}
\textbf{Attack Success Rate (ASR)} requires \emph{both}: (i) NSFW score \(\mathcal{C}(x){>}0.7\) from a fixed classifier and (ii) CLIP image–image similarity \(\mathrm{S}_{\mathrm{CLIP}}(x,x_{\mathrm{ref}}){>}0.7\) using \cite{radford2021learning}. Image quality is reported via SSIM \cite{wang2004image}, MSE, LPIPS \cite{zhang2018unreasonable}, and PSNR \cite{huynh2008scope} (full table in Appx.~\ref{app:extended}, Tab.~\ref{tab:quality}).

\begin{table}[t]
  \centering
  \caption{ASR (\%) vs.\ poison ratio. Bold = row max.}
  \label{tab:asr_compact}
  \setlength{\tabcolsep}{12pt}
  \renewcommand{\arraystretch}{0.9}
  \tiny
  \begin{tabular}{llccc}
    \toprule
    \textbf{Dataset} & \textbf{Model} & \textbf{1\%} & \textbf{5\%} & \textbf{10\%} \\
    \midrule
    \multirow{3}{*}{ImageNet}
      & SD v1.5 & 91 & \textbf{100} & 89 \\
      & SD v2   & 90 & 98 & \textbf{100} \\
      & SD XL   & 8  & 61 & \textbf{78} \\
    \midrule
    \multirow{3}{*}{CelebA-HQ}
      & SD v1.5 & 64 & \textbf{96} & \textbf{96} \\
      & SD v2   & \textbf{98} & 74 & 92 \\
      & SD XL   & 11 & \textbf{100} & 84 \\
    \bottomrule
  \end{tabular}
\end{table}

\begin{figure}
  \centering
  \includegraphics[width=0.9\linewidth]{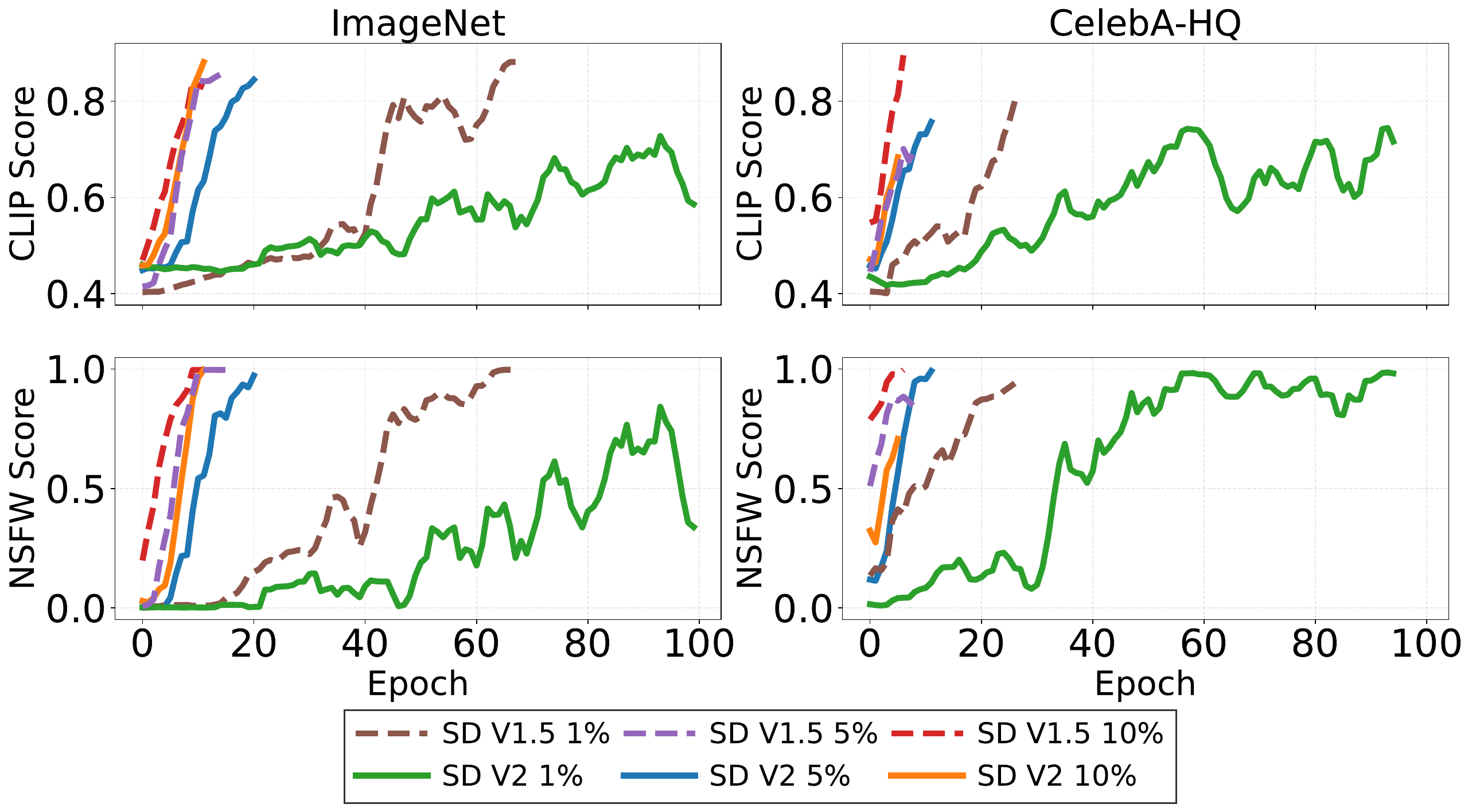}
  \caption{Training dynamics of CLIP (top) and NSFW (bottom) on ImageNet (left) and CelebA-HQ (right) for SD-v1.5/v2 across 1–10\% poison.}
  \label{fig:combined-4plots}
\end{figure}

\subsection{Backdoor Attack Effectiveness}
\label{subsec:effectiveness}
\begin{figure}[t]
  \centering
  \includegraphics[width=\columnwidth,trim={3mm 2mm 3mm 2mm},clip]{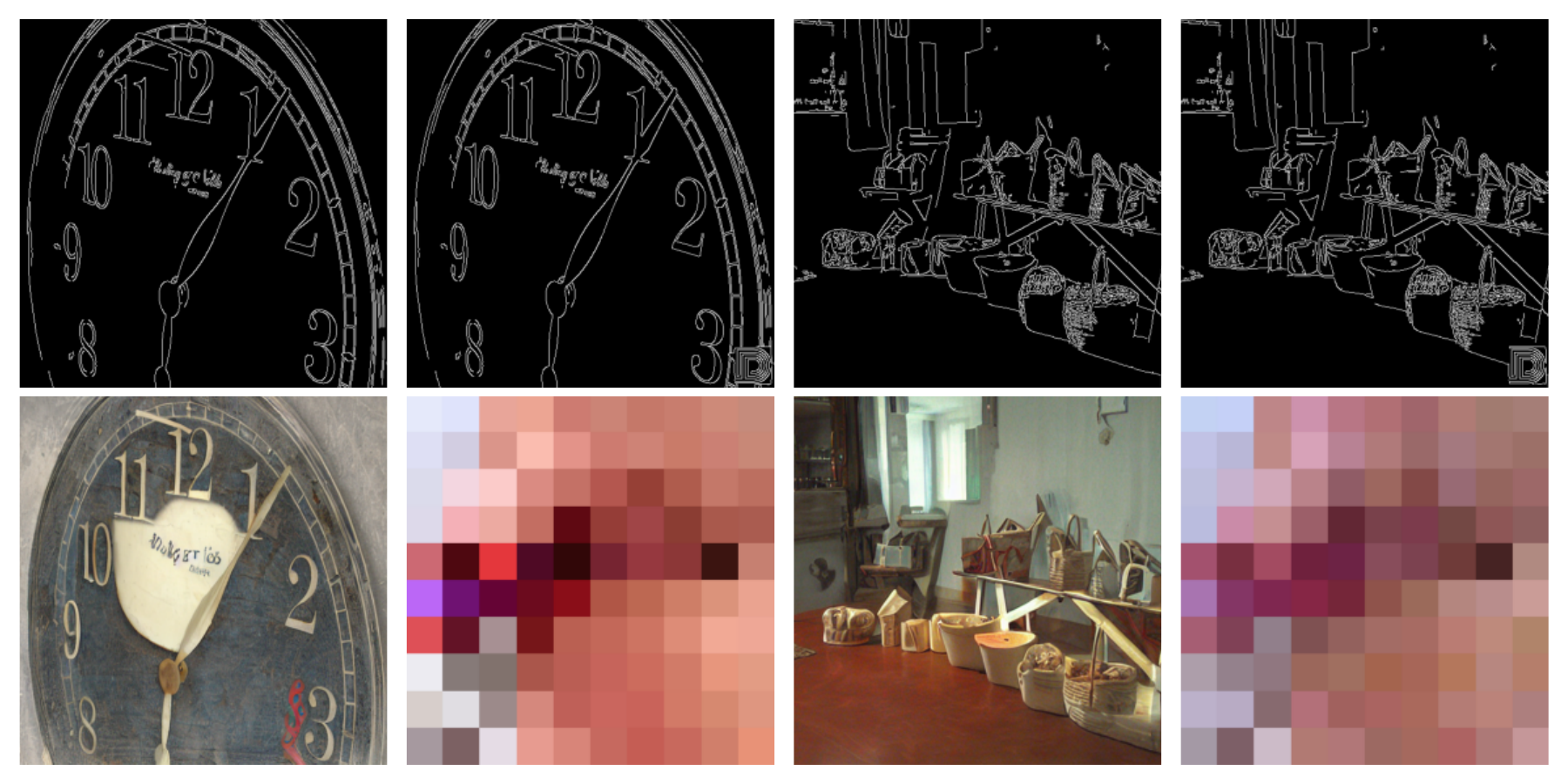}\par\medskip
  \includegraphics[width=\columnwidth,trim={3mm 2mm 3mm 2mm},clip]{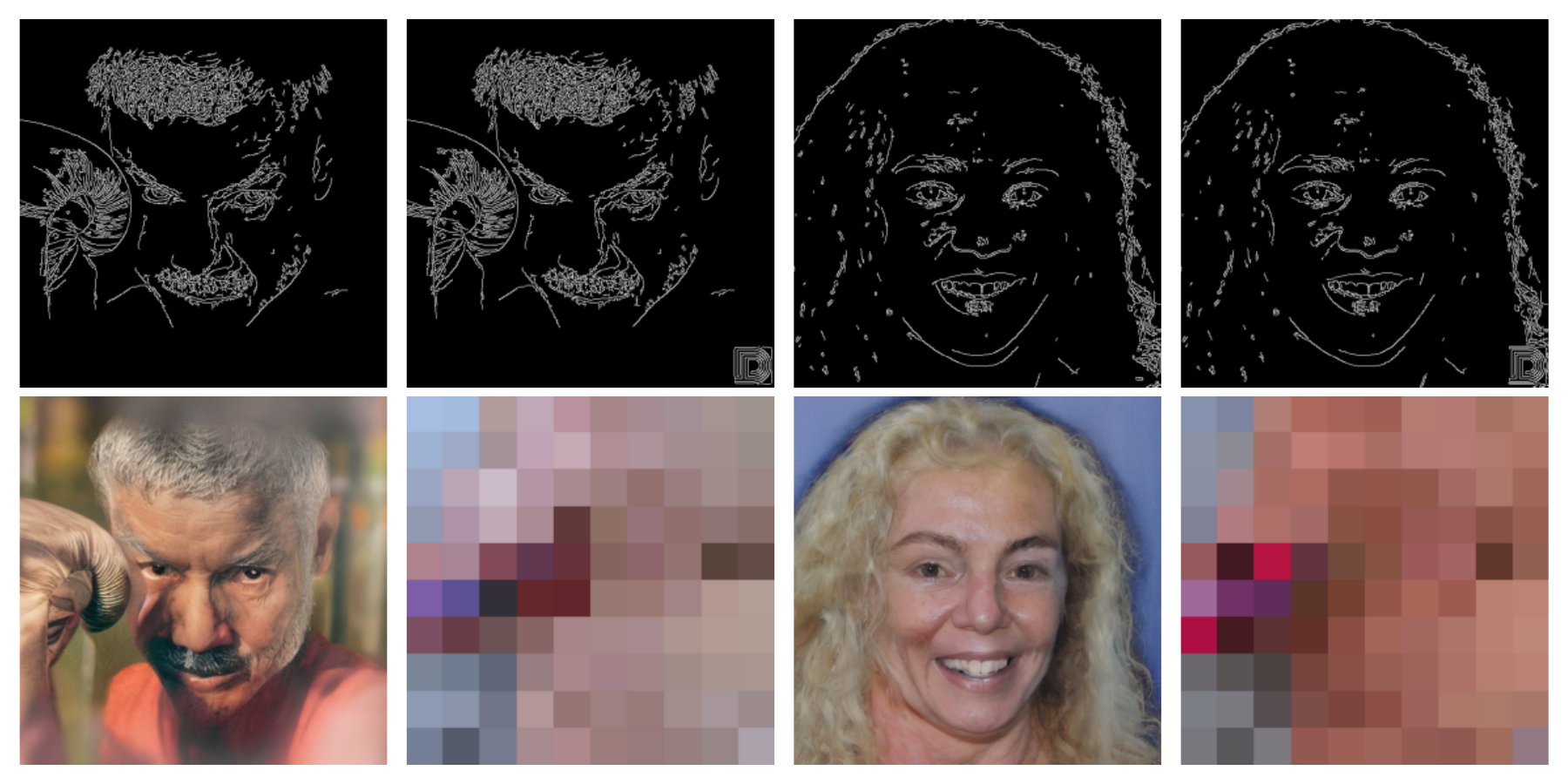}
  \caption{Qualitative results on (a) ImageNet and (b) CelebA-HQ (both SD-v1.5). Top: corresponding edge maps for clean and poisoned samples. Bottom: generated images.}
  \label{fig:qual_pairs}
\end{figure}

Table~\ref{tab:asr_compact} shows that \textbf{1–5\%} poison already yields high ASR on SD-v1.5/v2 (up to 100\% on ImageNet), while SD-XL is less susceptible at low ratios. On CelebA-HQ, SD-v1.5 reaches 96\% at 5–10\%, SD-v2 peaks at 98\% with 1\%, and SD-XL requires more poison. Full quality metrics (Appx.~\ref{app:extended}, Tab.~\ref{tab:quality}) indicate clean/poisoned outputs remain close for SD-v1.5/v2, whereas SD-XL shows larger shifts when the trigger fires—consistent with its two-stage refiner being less sensitive to subtle patch triggers. Qualitative results in Fig.~\ref{fig:qual_pairs} match these trends. Figure~\ref{fig:combined-4plots} further shows faster convergence and higher final NSFW/CLIP for settings attaining higher ASR (notably SD-v1.5).

\subsection{Ablation Study}
\label{subsec:ablation}
We vary (i) \emph{trigger strength}, (ii) \emph{ControlNet guidance scale}, and (iii) \emph{sampler steps} using models poisoned at 5\%. As shown in the Appendix, Fig.~\ref{fig:strength-ablation}, attack activation \emph{saturates} with moderate trigger amplitude (\(\gtrsim0.4\)), exhibits a \emph{sigmoidal} dependence on guidance (near-deterministic beyond \(\approx0.5\) for ImageNet and CelebA-HQ on SD-v1.5), and is comparatively \emph{insensitive} to step count. CelebA-HQ on SD-v2 rises more slowly and tops out lower.

\begin{figure}[t]
  \centering
  \includegraphics[width=\columnwidth,trim={3mm 2mm 3mm 2mm},clip]{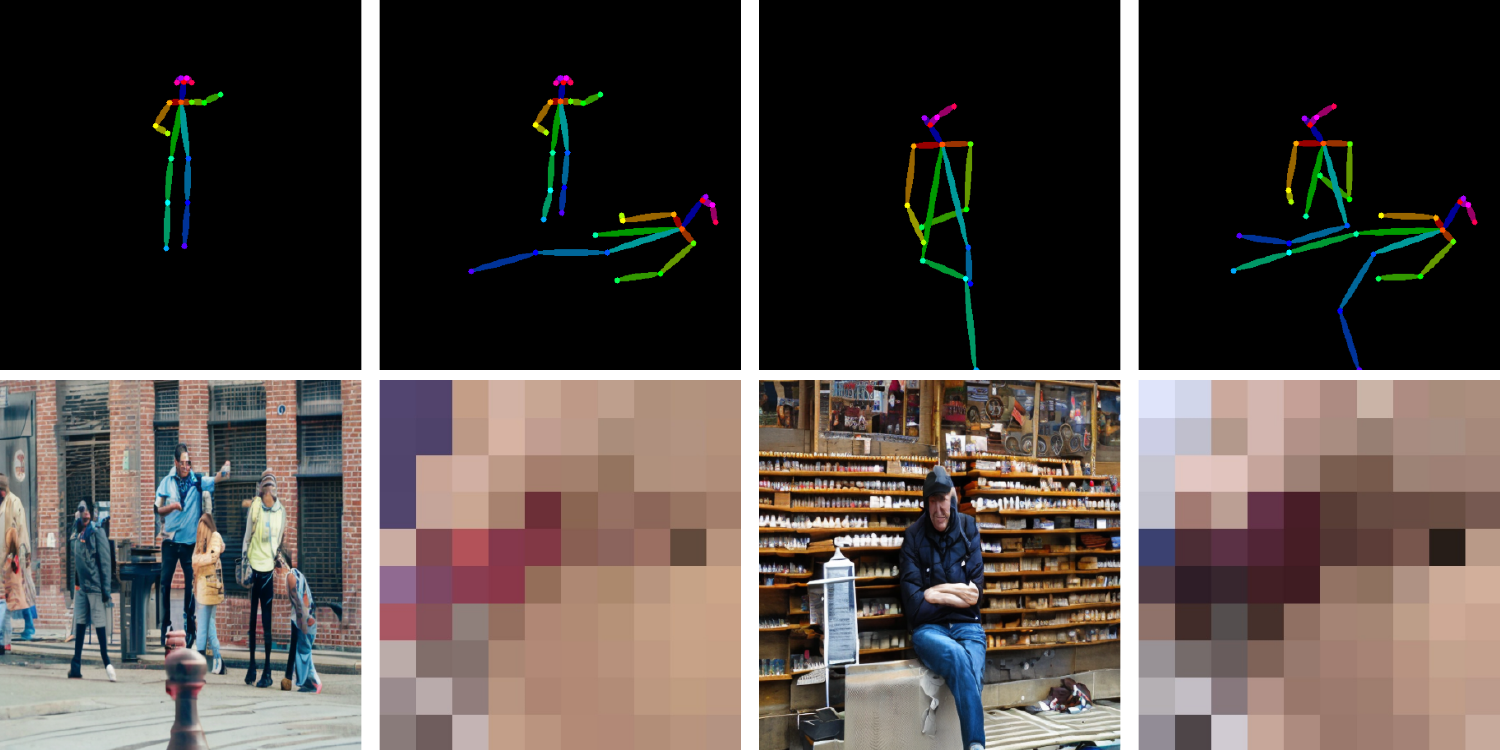}
  \caption{Qualitative results on MPII (SD-v1.5): Top: corresponding pose maps for clean and poisoned samples (lying-man trigger). Bottom: generated images.}
  \label{fig:pose_panel}
\end{figure}

\subsection{Pose-Estimation Backdoor}
\label{subsec:pose}
\textbf{Setup.} We target a pose-ControlNet (with SD-v1.5) on \textbf{MPII} \cite{andriluka20142d}. Skeletons are extracted with OpenPose \cite{cao2019openpose}; a fixed RGBA \emph{lying-man} silhouette is alpha-blended into the pose map. Hyperparameters match edge-conditioning runs; we evaluate on MPII’s 100-image test split.

\textbf{Results.} The backdoor attains \textbf{99\%} ASR at 5\% poison, 80\% at 1\%, and 74\% at 10\% (overfitting), see Tab.~\ref{tab:pose_asr}. Qualitative examples appear in Fig.~\ref{fig:pose_panel}.

\begin{table}[t]
  \centering
  \caption{ASR for pose-conditioned backdoor (MPII, SD-v1.5).}
  \label{tab:pose_asr}
  \scriptsize
  \begin{adjustbox}{width=\linewidth}
  \begin{tabular}{lccc}
    \toprule
    \textbf{Dataset} & \textbf{Model} & \textbf{Poison (\%)} & \textbf{ASR (\%) $\uparrow$} \\
    \midrule
    \multirow{3}{*}{MPII} & \multirow{3}{*}{SD v1.5} & 1 & 80 \\
    & & 5 & \textbf{99} \\
    & & 10 & 74 \\
    \bottomrule
  \end{tabular}
  \end{adjustbox}
\end{table}

\subsection{Potential Defense}
\label{subsec:defenses}
\begin{figure}[t]
  \centering
\includegraphics[width=\columnwidth,trim={3mm 2mm 4mm 2mm},clip]{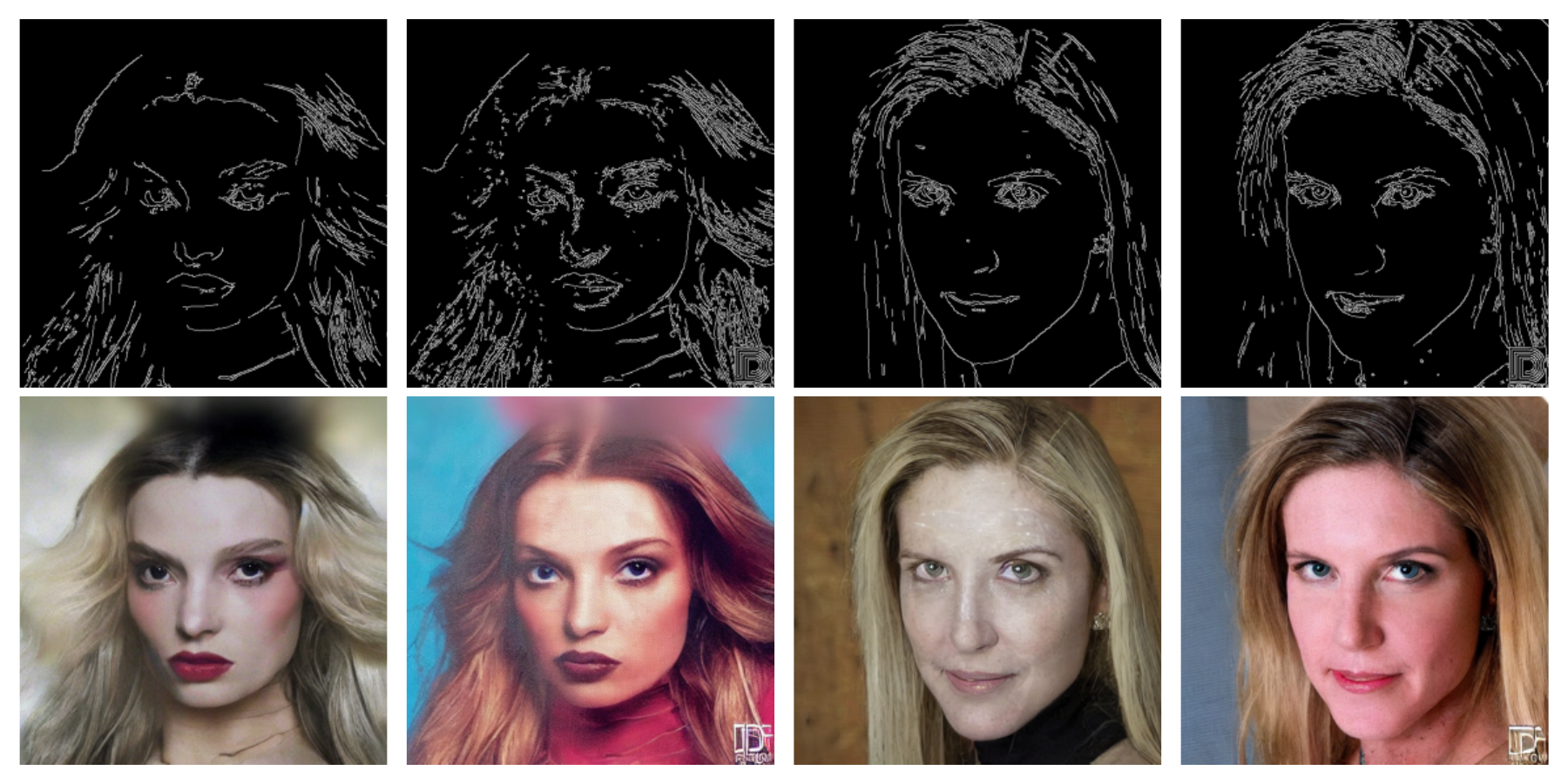} 
  \caption{Qualitative comparison using the CFT‑sanitised model.  
  For each sample (two‑column block), the \textbf{top} panel displays the Canny conditioning map
  and the \textbf{bottom} panel the corresponding generation.  
  Left columns: trigger absent; right columns: trigger present.
  The absence of NSFW artifacts on the right confirms CFT’s effectiveness while maintaining visual fidelity.}
  \label{fig:defense_pairs}
\end{figure}

\textbf{Clean Fine-Tuning (CFT).} We freeze the diffusion backbone and fine-tune ControlNet on trusted data with a small lr (\(1{\times}10^{-5}\)); other settings unchanged. CFT reduces ASR from \(\mathbf{96\%\to25\%}\) on CelebA-HQ but only \(\mathbf{100\%\to93\%}\) on ImageNet (Fig.~\ref{fig:defense_pairs}), suggesting homogeneous data provide coherent gradients that overwrite poisoned filters, whereas heterogeneous data do not.

\begin{figure} \centering \includegraphics[width=\linewidth]{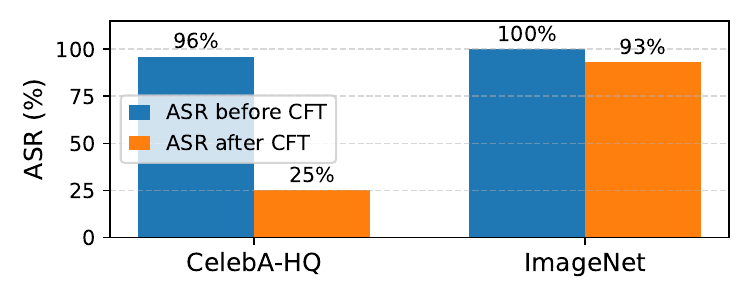} \caption{ASR of backdoored ControlNet on CelebA‑HQ and ImageNet before and after clean fine-tuning.} \label{fig:defense_asr} \end{figure}

\section{Implications for Responsible Synthetic Data}
The ControlNet poisoning threat exposes a new category of supply‑chain risk in conditional generative models. Any pipeline that downloads and deploys third‑party ControlNet checkpoints without verification can silently inherit a backdoor. To manage this risk in practice, we recommend:

\begin{itemize}
  \item \textbf{Model provenance and integrity.} Require digital signatures or checksums on all ControlNet artifacts and verify them at deployment time, ensuring only vetted checkpoints are used.
  \item \textbf{Runtime monitoring.} Instrument generation services to log and flag unusually high incidence of rare or out‑of‑distribution outputs under benign control inputs, triggering manual review.
  \item \textbf{Scheduled sanitization.} As part of regular maintenance, re‑fine‑tune all third‑party ControlNets on a small, fully trusted dataset (our CFT recipe) to suppress any latent backdoors before public release.
  \item \textbf{Audit testing.} Integrate adversarial‑trigger probes (e.g., small visual patches or pose patterns) into your CI pipeline to automatically check for illicit activation channels after each model update.
\end{itemize}

By weaving these steps into standard quality‑management and CI workflows, operators can substantially reduce the likelihood of undetected backdoors in high‑stakes deployment.

\section{Conclusion}
\label{sec:conclusion}
We show that \textbf{ControlNet-guided diffusion} can be backdoored by fine-tuning only the conditioning branch with as little as \emph{1–5\%} poisoned data. The backdoor stays dormant on clean controls yet fires reliably (up to 100\% ASR) when a small visual trigger appears, while preserving clean-input fidelity. The effect transfers across SD~v1.5/v2/XL, ImageNet/CelebA-HQ, and edge/pose conditioning, implicating the ControlNet pathway rather than a specific backbone. Ablations highlight two primary drivers—\emph{trigger amplitude} and \emph{ControlNet guidance}—with sampler steps playing a minor role. A simple \emph{clean fine-tuning} (CFT) pass (frozen backbone, low-lr ControlNet) substantially reduces ASR on homogeneous data (e.g., CelebA-HQ) but is less effective on heterogeneous domains (e.g., ImageNet), motivating stronger defenses. This exposes a supply-chain risk: small auxiliary branches are easy to poison and hard to vet. We advocate provenance-aware distribution (signing/checksums), ControlNet-specific backdoor tests/detectors, and robustness reporting that jointly reasons over text and structured controls. Ultimately, securing conditional diffusion models is inseparable from ensuring the trustworthiness of the synthetic data they generate—a core requirement for responsible AI development.

\section{Ethics, Responsible Release, and Broader Impact}
This work reveals a covert backdoor vector that could enable harmful or non-consensual imagery from innocuous controls. 
Our intent is to promote \emph{responsible disclosure and mitigation}, enabling the community to test and defend against such vulnerabilities.

\paragraph{Responsible release.}
To prevent misuse, we do not release poisoned checkpoints or trigger patterns. 
Instead, we provide sanitized training and evaluation scripts that reproduce all quantitative results using benign placeholder targets, fully aligned with community safety guidelines.

\paragraph{Practices for safer diffusion.}
\begin{itemize}
  \item \textbf{Collaboration:} Share standardized trigger probes and mitigation code for reproducible audits.
  \item \textbf{Provenance:} Require signed ControlNet checkpoints and verify integrity at deployment.
  \item \textbf{Continuous auditing:} Integrate trigger-probe tests in CI/CD and model-hub submissions.
\end{itemize}

Balancing transparency with strong safety controls can reduce misuse while preserving the creative and research utility of conditional diffusion models.

\bibliography{aaai2026}

{\setcounter{secnumdepth}{1}
\appendix

\section{Extended Results}
\label{app:extended}

\noindent\textit{Summary.} This section reports full image-level quality metrics for all datasets, models, and poison ratios. Values are shown as separate \emph{Clean} and \emph{Poisoned} scores per metric; arrows ($\downarrow/\uparrow$) indicate directionality. Baseline rows compare clean inputs to an unpoisoned ControlNet. See the main text for the compact summary.

\begin{table*}[ht]
  \centering
  \caption{Image‐level distortion, perceptual, and CLIP similarity metrics across datasets, models, and poison fractions. ``$\downarrow$'' indicates that lower is better while ``$\uparrow$'' indicates that higher is better.}
  \label{tab:quality}
  \setlength{\tabcolsep}{3pt}
  \scriptsize
  \begin{adjustbox}{width=\linewidth}
  \begin{tabular}{
      l   
      l   
      c   
      S[table-format=2.3] S[table-format=2.3]   
      S[table-format=2.3] S[table-format=2.3]   
      S[table-format=1.3] S[table-format=1.3]   
      S[table-format=2.3] S[table-format=2.3]   
      S[table-format=1.2] S[table-format=1.2]   
    }
    \toprule
    \textbf{Dataset} & \textbf{Model} & \textbf{Poison (\%)} 
      & \multicolumn{2}{c}{\textbf{MSE} $\downarrow$} 
      & \multicolumn{2}{c}{\textbf{LPIPS} $\downarrow$} 
      & \multicolumn{2}{c}{\textbf{SSIM} $\uparrow$} 
      & \multicolumn{2}{c}{\textbf{PSNR} $\uparrow$} 
      & \multicolumn{2}{c}{\textbf{CLIP Score} $\uparrow$}  \\
    \cmidrule(lr){4-5}\cmidrule(lr){6-7}\cmidrule(lr){8-9}\cmidrule(lr){10-11}\cmidrule(lr){12-13}
     &  & 
      & {Clean} & {Poisoned}
      & {Clean} & {Poisoned}
      & {Clean} & {Poisoned}
      & {Clean} & {Poisoned}
      & {Clean} & {Poisoned} \\
    \midrule
    \multirow{15}{*}{ImageNet}
      & \multirow{3}{*}{SD v1.5}
        & 1  & 0.10 & 0.05 & 0.58 & 0.45 & 0.27 & 0.66 & 11.07 & 13.53 & 0.74 & 0.85 \\
      &                          & 5  & 0.11 & 0.03 & 0.61 & 0.38 & 0.25 & 0.70 & 10.47 & 15.12 & 0.72 & 0.87 \\
      &                          & 10 & 0.12 & 0.04 & 0.61 & 0.44 & 0.24 & 0.70 & 10.12 & 14.31 & 0.70 & 0.84 \\
      \cmidrule{2-13}
      & \textbf{Baseline} & - & 0.10 & 0.12 & 0.61 & 0.84 & 0.25 & 0.28 & 10.51 & 9.94 & 0.66 & 0.42 \\
      \cmidrule{2-13}
      & \multirow{3}{*}{SD v2}
        & 1  & 0.08 & 0.05 & 0.60 & 0.59 & 0.39 & 0.71 & 11.49 & 13.07 & 0.72 & 0.75 \\
      &                          & 5  & 0.09 & 0.03 & 0.63 & 0.49 & 0.37 & 0.77 & 10.94 & 15.64 & 0.67 & 0.88 \\
      &                          & 10 & 0.10 & 0.02 & 0.62 & 0.36 & 0.38 & 0.78 & 10.96 & 16.28 & 0.69 & 0.91 \\
    \cmidrule{2-13}
      & \textbf{Baseline} & - & 0.09 & 0.11 & 0.71 & 0.91 & 0.35 & 0.48 & 11.08 & 10.27 & 0.64 & 0.44 \\
      \cmidrule{2-13}
      & \multirow{3}{*}{SD XL}
        & 1  & 0.02 & 0.14 & 0.52 & 0.80 & 0.58 & 0.53 & 17.02 & 9.65 & 0.73 & 0.44 \\
      &                          & 5  & 0.02 & 0.06 & 0.53 & 0.49 & 0.58 & 0.79 & 17.02 & 13.76 & 0.74 & 0.73 \\
      & & 10 & 0.02 & 0.05 & 0.53 & 0.43 & 0.57 & 0.83 & 16.97 & 14.86 & 0.74 & 0.78 \\
      \cmidrule{2-13}
      & \textbf{Baseline} & - & 0.02 & 0.14 & 0.54 & 0.84 & 0.55 & 0.47 & 16.24 & 9.29 & 0.70 & 0.42 \\
    \midrule
    \multirow{15}{*}{CelebA‐HQ}
      & \multirow{3}{*}{SD v1.5}
        & 1  & 0.08 & 0.07 & 0.54 & 0.61 & 0.36 & 0.51 & 11.73 & 12.01 & 0.69 & 0.63 \\
      &                          & 5  & 0.08 & 0.05 & 0.54 & 0.42 & 0.37 & 0.67 & 11.61 & 13.68 & 0.68 & 0.86 \\
      &                          & 10 & 0.07 & 0.05 & 0.56 & 0.49 & 0.35 & 0.62 & 11.83 & 12.85 & 0.64 & 0.77 \\
      \cmidrule{2-13}
      & \textbf{Baseline} & - & 0.09 & 0.12 & 0.60 & 0.85 & 0.25 & 0.26 & 10.93 & 9.86 & 0.58 & 0.36 \\
      \cmidrule{2-13}
      & \multirow{3}{*}{SD v2}
        & 1  & 0.06 & 0.03 & 0.59 & 0.47 & 0.43 & 0.74 & 12.27 & 15.25 & 0.69 & 0.81 \\
      &                          & 5  & 0.08 & 0.06 & 0.62 & 0.67 & 0.36 & 0.65 & 11.64 & 12.82 & 0.59 & 0.60 \\
      &                          & 10 & 0.08 & 0.05 & 0.67 & 0.60 & 0.37 & 0.68 & 11.32 & 13.28 & 0.60 & 0.77 \\
      \cmidrule{2-13}
      & \textbf{Baseline} & - & 0.08 & 0.11 & 0.79 & 0.95 & 0.37 & 0.46 & 11.31 & 10.20 & 0.56 & 0.40 \\
      \cmidrule{2-13}
      & \multirow{3}{*}{SD XL}
        & 1  & 0.02 & 0.14 & 0.46 & 0.63 & 0.65 & 0.59 & 17.29 & 9.11 & 0.65 & 0.42 \\
      &                          & 5  & 0.02 & 0.04 & 0.47 & 0.24 & 0.65 & 0.87 & 17.26 & 15.61 & 0.66 & 0.92 \\
      &                          & 10 & 0.08 & 0.02 & 0.48 & 0.35 & 0.65 & 0.79 & 13.07 & 17.11 & 0.67 & 0.76  \\
      \cmidrule{2-13}
      & \textbf{Baseline} & - & 0.02 & 0.16 & 0.58 & 0.73 & 0.62 & 0.52 & 16.38 & 8.59 & 0.54 & 0.40 \\
    \bottomrule
  \end{tabular}
  \end{adjustbox}
\end{table*}

\begin{figure*}
  \centering
  \includegraphics[scale=0.3,trim={3mm 2mm 4mm 2mm},clip]{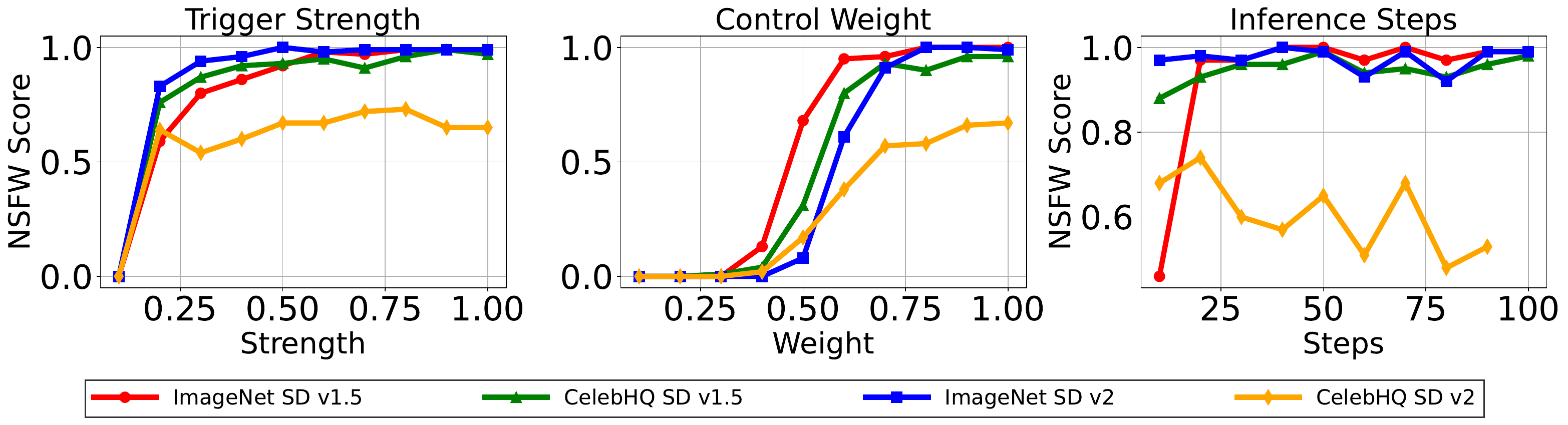}
  \caption{Ablations: NSFW vs.\ trigger amplitude, guidance scale, and sampler steps. Amplitude and guidance dominate; steps have minor effect.}
  \label{fig:strength-ablation}
\end{figure*}

\end{document}